\documentclass{article}
\usepackage{spconf,amsmath,graphicx}
\usepackage{algorithm}
\usepackage[noend]{algpseudocode}
\usepackage[sorting=none,backend=bibtex]{biblatex}
\usepackage{graphicx}
\usepackage{xcolor}
\usepackage{ragged2e}
\definecolor{ecgreen}{rgb}{0.5, 0.7, 0}
\definecolor{ecpurp}{rgb}{0.5, 0, 0.7}

\title{BOOSTING DICTIONARY LEARNING WITH ERROR CODES}
\name{Yigit Oktar$^	\dagger$, Mehmet Turkan$^\ddagger$}
\address{$^\dagger$Department of Computer Engineering\\
$^\ddagger$Department of Electrical and Electronics Engineering\\
Izmir University of Economics, Izmir, Turkey}
\begin{document}
\ninept
\maketitle

\begin{abstract}
In conventional sparse representations based dictionary learning algorithms, initial dictionaries are generally assumed to be proper representatives of the system at hand. However, this may not be the case, especially in some systems restricted to random initializations. Therefore, a supposedly optimal state-update based on such an improper model might lead to undesired effects that will be conveyed to successive iterations. In this paper, we propose a dictionary learning method which includes a general feedback process that codes the intermediate error left over from a less intensive initial learning attempt, and then adjusts sparse codes accordingly. Experimental observations show that such an additional step vastly improves rates of convergence in high-dimensional cases, also results in better converged states in the case of random initializations. Improvements also scale up with more lenient sparsity constraints. 
\end{abstract}


\section{INTRODUCTION}
\label{sec:intro}

One of the well-known properties of the Fourier transform and the wavelet transform is to exploit certain structures of the input signal and to represent these structures in a compact (or \textit{sparse}) manner. Sparse representations have thus become an active research topic while providing good performance in a large diversity of signal and image processing applications \cite{Elad_den06,Elad_den09,Peyre09, Mairal_resto08a,Mairal_resto08b, Elad_compr08,Peotta06, Fadili07,Mairal_disc08,Sapiro_nano08}.

Sparse representations consist in representing most or all information contained in a signal with a (linear) combination of a small number of \textit{atoms} carefully chosen from an over-complete (redundant) basis. This basis is generally referred to as a \textit{dictionary}. Such a dictionary is a collection of atoms whose number is much larger than the dimension of the signal space. Any signal then admits an infinite number of sparse representations and the sparsest such representation happens to have interesting properties for a number of signal and image processing tasks.

The most crucial question in sparse representations is the choice of the dictionary $\bf D$. One can realize a variety of pre-defined sets of waveforms, such as wavelets~\cite{MallatBook}, curvelets~\cite{DonohoBook}, contourlets~\cite{Do05}, shearlets~\cite{Labate05}, bandelets~\cite{Pennec05}. However, both the sparsity and the quality of the representation depend on how well the used dictionary is adapted to the data at hand. The problem of dictionary learning, or even simply finding adaptive ways to construct or to select relevant dictionaries, for sparse representations ---that goes far beyond using a few bases (like DCT, DFT, wavelets, or the others)--- has therefore become a key issue for further progress in this area.

Various dictionary learning algorithms have been proposed in the literature, e.g.,  \cite{MOD99, KSVD06, Lesage05, SOT08, Mairal_online10, EnganOnline, DoubleSparse, ITAD}. Most~of~these~methods~focus on $\ell_0$ or $\ell_1$ norm sparsity measures, potentially leading to simple formulations, hence to efficient techniques in practice. Method of Optimal Directions (MOD)~\cite{MOD99} is one of those successful methods of non-parametric dictionary learning. MOD builds upon the K-means process alternating between a sparse coding step and a least-squares optimization based update step of the dictionary. Although MOD is effective for low dimensional cases, because of pseudo-inverse computation, it becomes intractable in high dimensional problems. \mbox{K--SVD}~\cite{KSVD06} is a similar algorithm, alternating between sparse coding and dictionary update. The difference in dictionary update step is that, instead of updating $\bf D$ as a whole, K--SVD updates single dictionary atoms and corresponding sparse codes, using the rank-1 approximation by minimizing the approximation error when the corresponding atom has been isolated from the dictionary. Therefore, these iterative updates affect following ones, presumably accelerating convergence.

It is important to note that, both in MOD and K--SVD, sparse coding with $\ell_0$ norm constraint is assumed. Coding methods for such constraint include matching pursuit (MP) \cite{mallat1993matching} and orthogonal MP (OMP) \cite{pati1993orthogonal} algorithms in general. However, there is no theoretical restriction to use $\ell_1$ norm constraint instead. Therefore, in essence, both MOD and K--SVD define specific dictionary update procedures resulting in non-structural learned dictionaries, rather than defining the type of sparse coding constraint. On the other hand, certain dictionary learning algorithms are specifically based on coding with $\ell_1$ norm constraint. In this context, coefficient learning will be a more appropriate term than sparse coding when referring to calculation of sparse representation coefficients. Basis pursuit (BP)\cite{chen1994basis}, LARS~\cite{efron2004least}, FOCUSS \cite{gorodnitsky1995neuromagnetic} can be listed as examples. Gradient descent or Lagrange dual methods can then be chained as a dictionary update step \cite{lee2006efficient}. These alternative algorithms are proved to be more efficient especially in terms of convergence rates.

Accomplishments of such $\ell_1$ norm based algorithms over conventional $\ell_0$ ones are apparent. However, as noted before, $\ell_1$ norm is a relaxation on the original problem, so it does not guarantee an optimal solution in general. Such approach is especially problematic in low-dimensional cases. On the other hand, because of its non-convexity properties, $\ell_0$ norm approach leads to low convergence rates, being feasible only in low-dimensional cases.

A usual deficiency in sparse coding step is that, algorithms listed above assume proper dictionaries at each iteration. This is indeed very problematic, especially at initial stages of the learning process. In many situations, initial dictionary will not be a good representative of the optimal one. Therefore, ``optimal" coding done with such a dictionary, as targeted by both $\ell_0$ and $\ell_1$ norm coding schemes, will most likely result in sparse codes, which also are not good representatives of optimal state. As a result, the next dictionary will adopt this undesired property to a certain extent and convey it to successive learning steps. In this paper, we propose a generic modification to sparse coding or the coefficient learning step, with an error feedback process by coding an intermediate error and adjusting sparse codes accordingly in a less intensive learning attempt, hence leading to a faster convergence when compared to the conventional approaches. 

This paper is organized as follows. Section~\ref{sec:format} describes the details of the proposed dictionary learning algorithm. Section~\ref{sec:experiments} demonstrates the experimental setup and the obtained results. Finally, Section \ref{sec:conclusion} concludes this paper.

\section{Dictionary Learning with Error Codes}
\label{sec:format}

\subsection{Dictionary Learning}
\label{ssec:background}

The problem of dictionary learning for sparse representations can be formulated as a non-convex optimization problem in the form of the equation as below,
\begin{equation}
\begin{split}
\displaystyle{\mathop {\arg\min}\limits_{{\bf D}, \{{\bf x}_i\}}
\sum_{i=1}^{M}{\| {\bf y}_i-{\bf D}{\bf x}_i \|_2^2 }
\quad\textrm{subject to}\quad
\|{\bf x}_i\|_0 \leq k. }
\end{split}
\end{equation}
Here the set $\{{\bf y}_i\}, i=1...M,$ represents training samples, $\bf D$ is the dictionary to be learned and ${\bf x}_i$ is the sparse representation of the sample ${\bf y}_i$, $\forall i$. Parameter $k$ defines the sparsity constraint allowed for sparse coding during the learning process. Since Eqn. (1) poses a non-convex optimization problem through an $\ell_0$ norm constraint, solving this problem is thought to be NP-hard \cite{tillmann2015computational}. However, it is shown that for many high dimensional cases $\ell_1$ norm constraint is enough to ensure sparsest solution \cite{donoho2006most}. Note that $\ell_1$ norm constraint also turns the problem into a convex optimization one, which can then be solved via regular convex optimization tools. However, this should still be regarded as an approximation to the original problem.

In essence, traditional algorithms split Eqn. (1) into two approximate subproblems and alternate between these two simpler but convex optimization problems, namely \textit{sparse coding} and \textit{dictionary update} steps, to find a solution as follows,
\begin{equation}
\begin{split}
\displaystyle{\mathop{\arg\min}\limits_{\{{\bf x}_i\}}
\sum_{i=1}^{M}{\| {\bf y}_i-{\bf D}{\bf x}_i \|_2^2 } 
\quad\textrm{subject to}\quad
\|{\bf x}_i\|_0 \leq k,}
\end{split}
\end{equation}
\begin{equation}
\begin{split}
\displaystyle{
\mathop{\arg\min}\limits_{\bf D}
\sum_{i=1}^{M}{\| {\bf y}_i-{\bf D}{\bf x}^*_i\|_2^2 } }.
\end{split}
\end{equation}
Eqn. (2) corresponds to the sparse coding step where the dictionary $\bf D$ is assumed to be fixed. This subproblem can easily be solved with any pursuit algorithm. Eqn. (3) defines the dictionary update step that is performed with $\{{\bf x}^*_i\}$, i.e., sparse codes acquired from sparse coding. One direct way to solve this subproblem is the least-squares optimization as in MOD, such that ${\bf D}^*={\bf Y}{\bf X}^+$ where ${\bf Y} = \{{\bf y}_i\}_1^M$ and ${\bf X} = \{{\bf x}^*_i\}_1^M$ represent the training samples matrix and sparse codes matrix, respectively. ${\bf X}^+$ is the Moore-Penrose pseudo-inverse of $\bf X$. A full iteration is completed alternating between sparse coding and dictionary update, and this procedure is repeated until convergence.

\subsection{Introducing Error Feedback}
\label{ssec:learnfeed}
We propose a formulation that incorporates an intermediate error into the learning process. In the first stage, a regular sparse coding and dictionary update procedure is performed but with a sparsity level $m < k$ by solving Eqn. (4).
\begin{equation}
\begin{split}
\displaystyle{\mathop{\arg\min}\limits_{{\bf D}, \{{\bf a}_i\}}
\sum_{i=1}^{M}{\| {\bf y}_i-{\bf D}{\bf a}_i \|_2^2 }
\quad\textrm{subject to}\quad
\|{\bf a}_i\|_0 \leq m}
\end{split}
\end{equation}

\noindent Let us now denote $\{{\bf a}^*_i\}$ and ${\bf D}^*$ as the resulting sparse codes and the dictionary, respectively. The second stage involves sparse coding the approximation error ${\bf e}_i={\bf y}_i-{\bf D}^*{\bf a}^*_i$, $\forall i$, as
\begin{equation}
\begin{aligned}
\mathop{\arg\min}\limits_{\{{\bf b}_i\}}
\sum_{i=1}^{M}{\| {\bf e}_i - {\bf D}^*{\bf b}_i \|_2^2 }
\quad\textrm{s.t.}\quad
\|{\bf b}_i\|_0 \leq k-m.  \\
\end{aligned}
\end{equation}
After acquiring $\{{{\bf b}^*_i}\}$ in Eqn. (5), current-state sparse codes can further be updated as ${{\bf a}^*_i + {\bf b}^*_i}$. This step basically corresponds to some sort of feedback logic, where the first approximation is tested and then its deviation is sparse coded to be incorporated into actual codes. Note here that the original sparsity constraint still holds since ${\|{\bf a}^*_i +{\bf b}^*_i\|_0 \leq k}$. In the last stage, a final dictionary update is performed as in Eqn.~(6) and an iteration is completed,
\begin{equation}
\begin{split}
\displaystyle{
\mathop{\arg\min}\limits_{\bf D}
\sum_{i=1}^{M}{\| {\bf y}_i-{\bf D}({\bf a}^*_i+{\bf b}^*_i)\|_2^2 } }.
\end{split}
\end{equation}

\section{EXPERIMENTAL RESULTS}
\label{sec:experiments}

Two variants of the proposed scheme have been tested experimentally, namely EcMOD in Algorithm 1 and EcMOD+ in Algorithm 2. EcMOD includes the methodology that is defined in Section 2, and EcMOD+ includes a regular least-squares dictionary update (MOD) at the end of each iteration. OMP is used for sparse coding. 

\begin{figure*}
  \centering \includegraphics[scale=0.32]{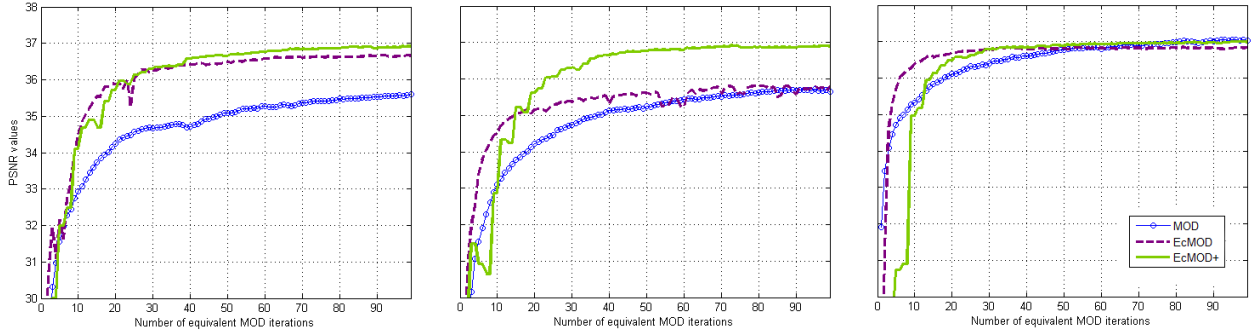}
  \caption{Results of dictionary learning performed on the \textit{Barbara} image. Dictionary size $64\times 256$, $k=8$ for both learning and testing. Each column represents a different initial dictionary. First and middle columns are randomly initialized. Last column is of DCT initialization. Figures depict PSNR (dB) performance values versus iteration number.}
\end{figure*}

\begin{algorithm}[t]
\label{ecmod}
\begin{algorithmic}[1]
\caption{EcMOD algorithm pseudocode.}
\Function{EcMOD}{${\bf D}, {\bf Y} , m, k$}
\While {$not \hspace{2pt} converged$}
\State ${\bf A} \gets$ OMP(${\bf D}, {\bf Y}, m$)
\State ${\bf D} \gets$ ${\bf Y}{\bf A}^+$
\State ${\bf E} \gets {\bf Y}-{\bf D}{\bf A}$
\State ${\bf B} \gets$ OMP(${\bf D}, {\bf E}, k-m$)
\State ${\bf D} \gets$ ${\bf Y}({\bf A}+{\bf B})^+$
\EndWhile
\Return $\bf D$
\EndFunction
\end{algorithmic}
\end{algorithm}

\begin{algorithm} [t]
\begin{algorithmic}[2]
\caption{EcMOD+ algorithm pseudocode.}
\Function{EcMOD+}{${\bf D}, {\bf Y}, m, k$}
\While {$not \hspace{2pt} converged$}
\State ${\bf D} \gets$ EcMOD(${\bf D}, {\bf Y}, m, k$)
\State ${\bf X} \gets$ OMP(${\bf D}, {\bf Y}, k$)
\State ${\bf D} \gets$ ${\bf Y}{\bf X}^+$
\EndWhile
\Return $\bf D$
\EndFunction
\end{algorithmic}
\end{algorithm}

Two experimental setups have been performed, corresponding to low and high dimensional cases respectively. In the first setup, $8\times 8$ distinct patches were extracted from the \textit{Barbara} image of size $512\times 512$, resulting in $4096$ image patches. Dictionary size was accordingly chosen as $64\times 256$. Sparsity constraint $k$ and additional sparsity parameter $m$ were chosen as $8$ and $4$ respectively, so $m$ being equal to $k/2$. Results corresponding to this setup are presented in Figure 1, Figure 2 and Table 1.

In error coded schemes, as a consequence of not directly coding with sparsity $k$, final codes ${\bf a}^*_i + {\bf b}^*_i$ may not necessarily be optimal~for~$k$. However, as there are two coding steps with lesser sparsity constraints and a summation, codes have a higher chance of being optimal for most of the sparsity levels less than $k$. As a result, converged dictionary is a better representative of such sparsity levels, as observable in the results in Table 1. Error coded schemes consistently perform better in sparser cases.

Not targeting a sparsity level $k$ directly leads to a possibility of converged dictionary to be suboptimal for that given $k$. However, this drawback can be worked around by chaining a conventional step that targets an exact sparsity level $k$. Referred to as EcMOD+ algorithm, experiments with this further modified method show that, such architecture possesses optimality for $k$ sparsity and also better performance for the cases where sparsity level is less than $k$. This phenomenon is apparent in Figure 1, where learning and testing k chosen both as 8. Performance of EcMOD is not consistent as it performs much like MOD in second random initialization case. Whereas, EcMOD+ consistently performs well.

\begin{figure}[t]

\begin{minipage}[b]{1.0\linewidth}
  \centering
  \centerline{\includegraphics[scale=0.605]{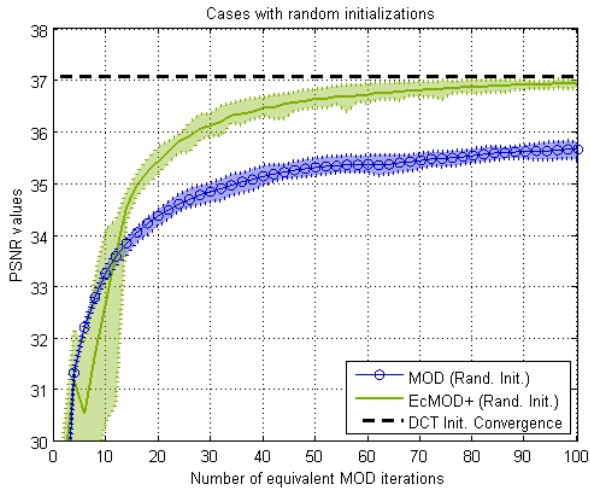}}
 
\end{minipage}

\caption{Results for ten cases of uniformly random initial dictionaries. Lower and upper lines of each method correspond to minimum and maximum PNSR (dB) values attained among all ten cases. Middle lines represent average values.}
\label{fig:resrand}
\end{figure}

In a more extensive manner, Figure 2 compares the performance of MOD and EcMOD+ in the case of ten different randomly initialized dictionaries. DCT convergence is supplied as a baseline. This figure represents superiority of this error coding scheme over conventional coding in the case of random initializations. "Optimal" coding with an improper random dictionary within initial stages hamper the final convergence state as observable in the case of MOD. Although not targeting optimal codes, error coded scheme EcMOD+ converges to DCT result in all 10 random initialization cases. This is possible because, in each step from the beginning, dictionary passes through a less intensive validation, in the expense of acquiring optimal codes. There is no total superiority in DCT initialization case as seen in Table 1. However, superiority can be achieved with more complex error coding schemes.


\begin{table}[t]

\centering
\begin{tabular}{ l r c c c c}
\hline

  & \quad \quad \quad \quad  & $k=2$ & $k=5$ & $k=10$ & $k=20$\\

\hline
\textbf{Rand. 1.} \\
\textcolor{blue}{MOD} & & 26.33 & 32.08 & 36.82 & 40.43 \\
\textcolor{ecpurp}{EcMOD} & & \textbf{26.91} & \textbf{33.90} & 37.87 & 41.72\\
\textcolor{ecgreen}{EcMOD+}& & 26.89 & \textbf{33.90} & \textbf{38.10} & \textbf{41.86} \\
\hline

\textbf{Rand. 2.}  \\
\textcolor{blue}{MOD} & & 26.61 & 32.20 & 36.89 & 40.68\\
\textcolor{ecpurp}{EcMOD} & & \textbf{27.92} & 33.36 & 36.88 & 40.25\\
\textcolor{ecgreen}{EcMOD+} & & 26.73 & \textbf{33.87} & \textbf{38.08} & \textbf{41.83}\\
\hline

\textbf{DCT Init.}  \\
\textcolor{blue}{MOD}& & 26.22 & 33.02 & \textbf{38.40} & \textbf{42.82} \\
\textcolor{ecpurp}{EcMOD} & & \textbf{26.95} & \textbf{34.05} & 38.06 & 41.98\\
\textcolor{ecgreen}{EcMOD+}& & 26.94 & 33.96 & 38.19 & 42.07\\
\hline

\end{tabular}
\par
\bigskip

\justifying
\noindent
\textbf{Table 1.} Average approximation PSNR (dB) performance values of learnt dictionaries as in Figure 1 and Figure 2. $k$ represents the sparsity used for testing.
\end{table}

\begin{figure}[!t]

\begin{minipage}[b]{1.0\linewidth}
  \centering
  \centerline{\includegraphics[scale=0.64]{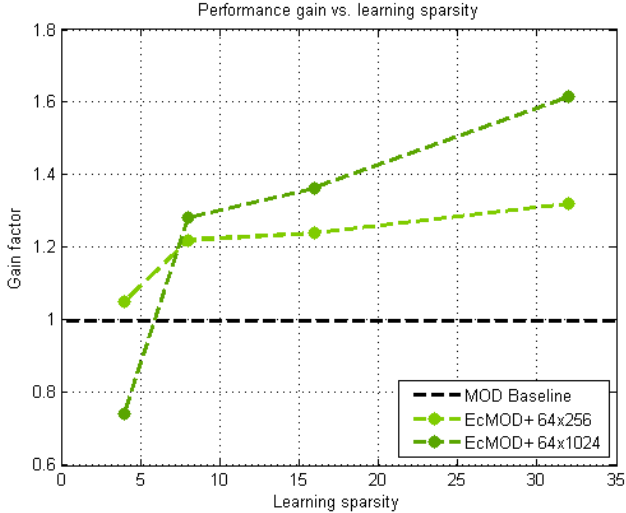}}
 
\end{minipage}

\caption{Performance gain factor for different sparsity levels, in the case of a relatively small and a large dictionary.}

\end{figure}

\begin{figure}[!t]

\begin{minipage}[b]{1.0\linewidth}
  \centering
  \centerline{\includegraphics[scale=0.64]{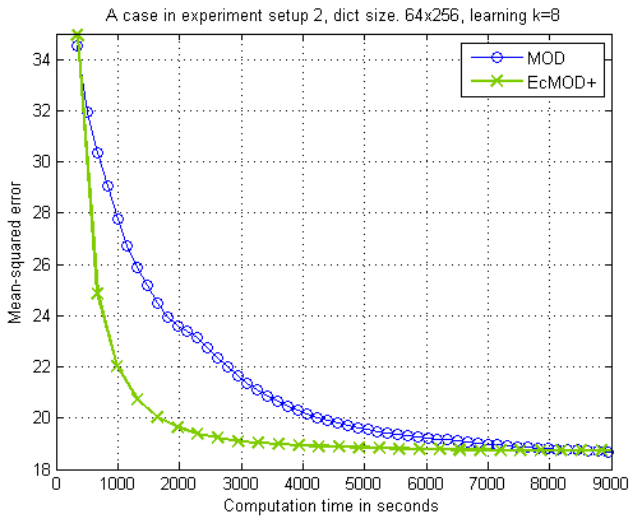}}
 
\end{minipage}

\caption{The convergence performance for dictionary size 64 x 256
and k as 8 for both learning and testing.}

\end{figure}

Finally, regarding overall sparsity levels within error codes and its evolution during learning process, the proposed method presents interesting trends. In conventional sparse coding (targeting the sparsity $k$), as approximation threshold is kept very strict in general, sparse codes end up with using all $k$ supports. Therefore, in methods such as MOD and K--SVD, codes consistently have $k$ supports even starting from the initial iteration. In the proposed method, during error coding, selection of previously selected supports is frequent. This is especially observable during initial iterations. Near maximum support counts are gradually reached as the system converges, but not necessarily reaching exact maximum.

In the second set of experiments, EcMOD+ scheme has been tested with all possible $255025$ image patches extracted with a full coverage of sliding window algorithm with a window \mbox{size~of~$8\times 8$}. Combinations of sparsity of $4$, $8$, $16$, and $32$ against small and large dictionaries were tested. DCT was used as initial dictionary in all cases. Note that, in DCT initialization cases, there is only an advantage of faster convergence rates but not of better converged states, at least for this error coding scheme. Superior converged states with more complex schemes have been achieved, but they are omitted here because of the space limitation. 

Experimental results with the second setup are summarized in Figure 3 and Figure 4. In Figure 3, performance ratios were calculated relative to the MOD algorithm in terms of mean-squared error to estimate a performance gain factor for each sparsity level, approximately at fifth equivalent iteration, for an approximate convergence rate analysis. Gains for large dictionary in the case of learning with lenient sparsity levels are more striking, but stricter sparsity constraints cause substandard performance. Overall, the performance in this case is promising as it signals to scaling with input size. Smaller sized dictionary safely performs above standard. Figure 4 depicts the convergence plot for dictionary size 64 x 256 and k as 8 for
both learning and testing. Significant gain in convergence rate is observable
with the error coded scheme in this high-dimensional setup.
Note that, mean-squared error is given as measure since sliding window
patches were used. Finally, Figure 5 compares atoms that lie
within similar frequency domains, learnt with MOD and EcMOD+.
Note here the well-defined structure of EcMOD+.

\begin{figure}[!t]

\begin{minipage}[b]{1.0\linewidth}
  \centering
  \centerline{\includegraphics[scale=0.37]{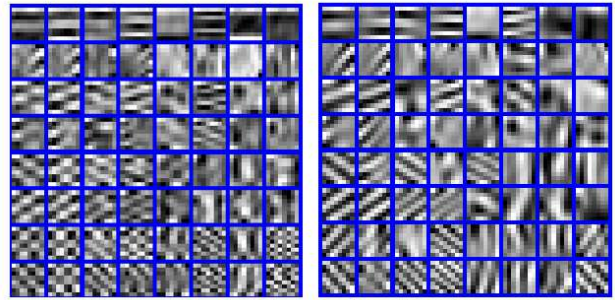}}
 
\end{minipage}

\caption{A frequency subdomain of learnt dictionaries formed with same processing time. MOD on the left, EcMOD+ on the right. }

\end{figure}

\section{CONCLUSION}
\label{sec:conclusion}
MOD, by itself is a greedy algorithm that targets optimality one task at a time.  Tasks are considered as isolated from each other, even within the same iteration. This results in MOD being a rather short-sighted method which fails at tasks that require a broader perspective of the system.

In this paper, we presented a method in which sparse coding and dictionary update steps are intertwined through intermediate error codes. Note that there could be other ways to accomplish this. Another way could be to add sparse codes of two successive iterations and perform a dictionary update based on this accumulated code, without even introducing error codes. As an analogy, MOD can be considered as a single-step numerical method, where as the example given would be a multi-step one. Our method can be considered as a multi-step approach that utilizes a half-step.

To summarize, our framework is generic enough to be included in many forms of learning-based approaches. In essence, our scheme includes an initial attempt of learning with less computational and spatial requirement than originally allocated. This corresponds to a single iteration of MOD performed with $m < k$. In this way, a feedback can be acquired that reflects the
congruence of the model and the data at hand, so that current state can properly be adjusted before the final model update, which consumes the remaining resources. 

This approach will be most beneficial for systems that are restricted to random initializations (as apparent in Figure 1, 2 and Table 1). A random initial model is most likely to be an improper representative of a specific system. Therefore, an update based on this model, no matter how intensive it is, will result in an undesired state. In fact, an optimal update based on this improper model could be more impairing than a suboptimal one in this regard.

As a concluding remark, readers should bear in mind that this work is based on a pragmatic perspective. Although satisfactory improvement has been observed, a more rigorous theoretical approach can lead to certain variations built on top of this framework that will be far more fruitful.


\clearpage\clearpage

\ninept


\printbibliography

\end{document}